%% file: main.tex
\documentclass[11pt,a4paper]{article}

\usepackage[utf8]{inputenc}
\usepackage{multicol}

\usepackage{subfigure}
\usepackage[round,authoryear]{natbib}

\usepackage{bm}
\usepackage{amsmath}
\usepackage{boxedminipage}

\usepackage{listings}
\usepackage{color}
\definecolor{codegreen}{rgb}{0,0.6,0}
\definecolor{codegray}{rgb}{0.5,0.5,0.5}
\definecolor{codepurple}{rgb}{0.58,0,0.82}
\definecolor{backcolour}{rgb}{0.95,0.95,0.92}
\definecolor{codered}{rgb}{0.5,0,0}

\lstdefinestyle{mystyle}{
    backgroundcolor=\color{backcolour},   
    commentstyle=\color{codered},
    keywordstyle=\color{magenta},
    numberstyle=\tiny\color{codegray},
    stringstyle=\color{codegreen},
    breakatwhitespace=false,         
    breaklines=true,                 
    captionpos=b,                    
    keepspaces=true,                 
    numbers=left,                    
    numbersep=5pt,                  
    showspaces=false,                
    showstringspaces=false,
    showtabs=false,                  
    tabsize=2,
    basicstyle=\ttfamily\footnotesize
}
 
\usepackage{ifpdf}

\usepackage[T1]{fontenc}
\usepackage{fancyhdr}   % pour les en-tete et pieds de pages
\usepackage{makeidx}

\usepackage{lipsum}
\usepackage{verbatim}

\ifpdf
\usepackage[pdftex]{graphicx}
\else
\usepackage{graphicx}
\fi
\usepackage{rotating}
\usepackage{blindtext}
\usepackage[includemp=false, marginparwidth=1.8cm, hmargin=3.5cm, vmargin=3cm]{geometry}
\ifnum\pdfoutput>0			% Include hyperref if outputting pdf
\usepackage[pdftex, 			% Paramétrage de la navigation
pagebackref = false, 			% Ajoute les liens vers les pages dans la bibliographie
bookmarks = true, 			% Signets
bookmarksnumbered = false, 	% Signets numérotés
hyperindex = true,			% Hypertexte dans l'index
linktocpage = true,			% Le lien porte sur le numéro, pas sur le titre, dans la toc
pdfpagemode = UseNone, 		% Full Screen ou None
pdfstartview = Fit, 			% La page prend toute la largeur
pdfpagelayout = OneColumn,% Vue par page
colorlinks = true, 			%false, % Liens en couleur
urlcolor = blue, 				% Couleur des liens externes
citecolor = blue,
linkcolor = blue,
pdfborder = {0 0 0} 			% Style de bordure : ici, pas de bordure
]{hyperref} 					% Utilisation de HyperTeX
\fi

\ifpdf
\DeclareGraphicsExtensions{.png, .pdf, .jpg, .tif}
\else
\DeclareGraphicsExtensions{.eps, .jpg}
\fi

\input{macros.tex}

\graphicspath{{figures/}}

\title{{\bf \huge DeepMind Lab}}
\author{ 
\small Charles Beattie,
Joel Z. Leibo,
Denis Teplyashin,
Tom Ward,\\
\small Marcus Wainwright,
Heinrich Küttler,
Andrew Lefrancq,\\
\small Simon Green,
Víctor Valdés,
Amir Sadik,
Julian Schrittwieser,\\
\small Keith Anderson,
Sarah York,
Max Cant,
Adam Cain,\\
\small Adrian Bolton,
Stephen Gaffney,
Helen King,\\
\small Demis Hassabis,
Shane Legg
and Stig Petersen
}
\date{\small \today}

\begin{document}

\maketitle

\begin{abstract}
DeepMind Lab is a first-person 3D game platform designed for research and development of general artificial intelligence and machine learning systems. DeepMind Lab can be used to study how autonomous artificial agents may learn complex tasks in large, partially observed, and visually diverse worlds. DeepMind Lab has a simple and flexible API enabling creative task-designs and novel AI-designs to be explored and quickly iterated upon. It is powered by a fast and widely recognised game engine, and tailored for effective use by the research community.
\end{abstract}

%\begin{center}
%\includegraphics[height=20ex]{./figs/logo}
%\end{center}

%Comment the following if you don't want a table of contents
%\tableofcontents

%------------------------------------------

%%%%%%%%%%%%%%%%%%%%%%%%%%%%%%%%%%%%%%%%%%%%%%%%%%%%%%%%%
\section*{Introduction}

General intelligence measures an agent's ability to achieve goals in a wide range of environments \citep{legg2007universal}.  The only known examples of general-purpose intelligence arose from a combination of evolution, development, and learning, grounded in the physics of the real world and the sensory apparatus of animals. An unknown, but potentially large, fraction of animal and human intelligence is a direct consequence of the  perceptual and physical richness of our environment, and is unlikely to arise without it \citep[e.g.][]{locke1690essay, hume1739treatise}. One option is to directly study embodied intelligence in the real world itself using robots \citep[e.g.][]{brooks1990elephants, metta2008icub}. However, progress on that front will always be hindered by the too-slow passing of real time and the expense of the physical hardware involved. Realistic \emph{virtual} worlds on the other hand, if they are sufficiently detailed, can get the best of both, combining perceptual and physical near-realism with the speed and flexibility of software.

Previous efforts to construct realistic virtual worlds as platforms for AI research have been stymied by the considerable engineering involved. To fill the gap, we present DeepMind Lab. DeepMind Lab is a first-person 3D game platform  built on top of id software’s Quake III Arena \citep{QuakeThree} engine. The world is rendered with rich science fiction-style visuals.  Actions are to look around and move in 3D. Example tasks include navigation in mazes, collecting fruit, traversing dangerous passages and avoiding falling off cliffs, bouncing through space using launch pads to move between platforms, laser tag, quickly learning and remembering random procedurally generated environments, and tasks inspired by Neuroscience experiments. DeepMind Lab is already a major research platform within DeepMind. In particular, it has been used to develop asynchronous methods for reinforcement learning \citep{mnih2016asynchronous}, unsupervised auxiliary tasks \citep{jaderberg2016reinforcement}, and to study navigation \citep{mirowski2016learning}.

DeepMind Lab may be compared to other game-based AI research platforms emphasising pixels-to-actions autonomous learning agents. The Arcade Learning Environment (Atari) \citep{bellemare2012arcade}, which we have used extensively at DeepMind, is neither 3D nor first-person. Among 3D platforms for AI research, DeepMind Lab is comparable to others like VizDoom \citep{kempka2016vizdoom} and  Minecraft \citep{johnson2016malmo, tessler2016deep}. However, it pushes the envelope beyond what is possible in those platforms. In comparison, DeepMind Lab has considerably richer visuals and more naturalistic physics. The action space allows for fine-grained pointing in a fully 3D world. Compared to VizDoom, DeepMind Lab is more removed from its origin in a first-person shooter genre video game. This work is different and complementary to other recent projects which run as plugins to access internal content in the Unreal engine \citep{qiu2016unrealcv, lerer2016learning}. Any of these systems can be used to generate static datasets for computer vision as described e.g., in \cite{mahendran2016researchdoom, richter2016playing}.

Artificial general intelligence (AGI) research in DeepMind Lab emphasises 3D vision from raw pixel inputs, first-person (egocentric) viewpoints, fine motor dexterity, navigation, planning, strategy, time, and fully autonomous agents that must learn for themselves what tasks to perform by exploration of their environment. All these factors make learning difficult. Each are considered frontier research questions on their own. Putting them all together in one platform, as we have, is a significant challenge for the field.

%%%%%%%%%%%%%%%%%%%%%%%%%%%%%%%%%%%%%%%%%%%%%%%%%%%%%%%%%

\section*{DeepMind Lab Research Platform}

DeepMind Lab is built on top of id software’s Quake III Arena \citep{QuakeThree} engine using the ioquake3 \citep{nussel2016ioquake3} version of the codebase, which is actively maintained by enthusiasts in the open source community. DeepMind Lab also includes tools from q3map2 \citep{gtkradiant} and bspc \citep{bspc} for level generation. The bot scripts are based on code from the OpenArena \citep{OpenArena} project.

\subsection*{Tailored for machine learning}

A custom set of assets were created to give the platform a unique and stylised look and feel, with a focus on rich visuals tailored for machine learning.

A reinforcement learning API has been built on top of the game engine, providing agents with complex observations and accepting a rich set of actions.

The interaction with the platform is lock-stepped, with the engine stepped forward one simulation step (or multiple with repeated actions, if desired) at a time, according to a user-specified frame rate. Thus, the game is effectively paused after an observation is provided until an agent provides the next action(s) to take.

\subsection*{Observations}
At each step, the engine provides reward, pixel-based observations and, optionally, velocity information (figure \ref{fig:observations}):

\begin{figure}[t]
\centering
\includegraphics[scale = 0.22]{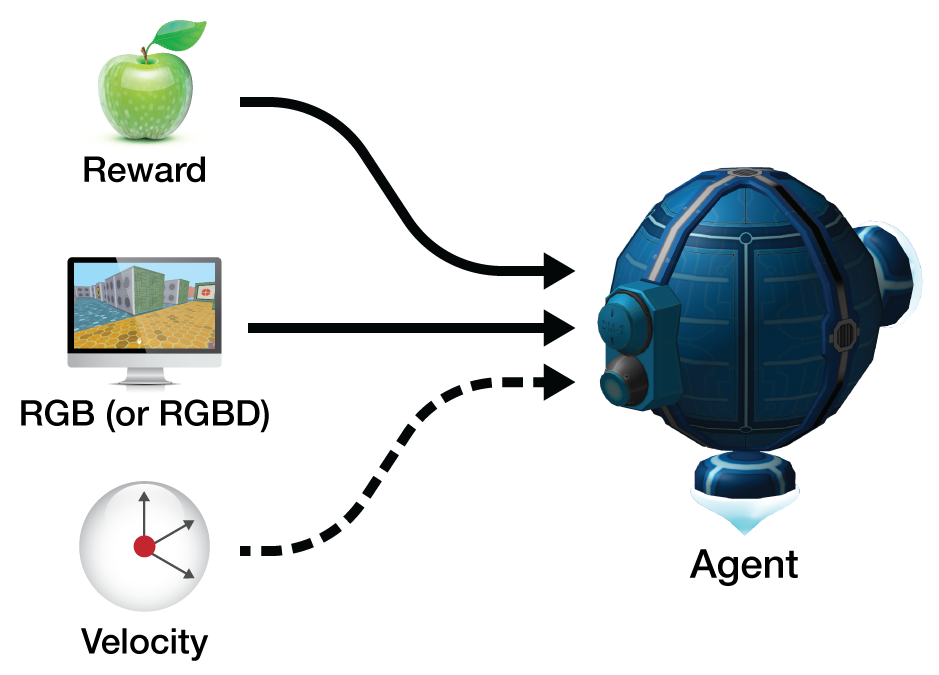}
  \caption{Observations available to the agent. In our experience, reward and pixels are sufficient to train an agent, whereas depth and velocity information can be useful for further analysis.
  \label{fig:observations} }
\end{figure}

\begin{enumerate}
\item The reward signal is a scalar value that is effectively the score of each level.
    
\item The platform provides access to the raw pixels as rendered by the game engine from the player’s first-person perspective, formatted as RGB pixels. There is also an RGBD format, which additionally exposes per-pixel depth values, mimicking the range sensors used in robotics and biological stereo-vision.

\item For certain research applications the agent’s translational and angular velocities may be useful. These are exposed as two separate three-dimensional vectors.
\end{enumerate}

\subsection*{Actions}
\begin{figure}[t]
\centering
\includegraphics[scale = 0.22]{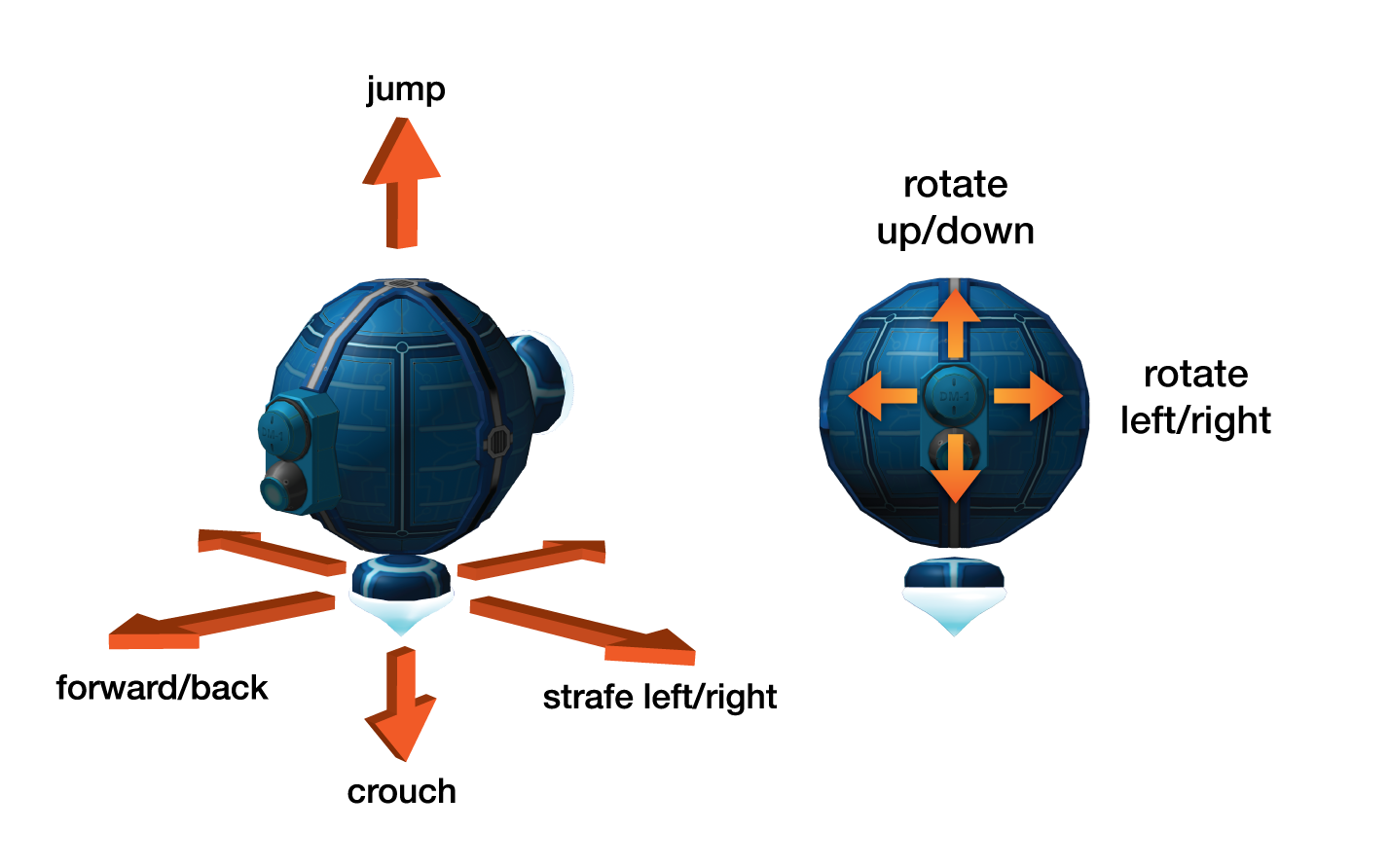}
  \caption{The action space includes movement in three dimensions and look direction around two axes.
  \label{fig:actions} }
\end{figure}

Agents can provide multiple simultaneous actions to control movement (forward/back, strafe left/right, crouch, jump), looking (up/down, left/right) and tagging (in laser tag levels with opponent bots), as illustrated in figure \ref{fig:actions}.

\subsection*{Example levels}
\begin{figure}[t]
\centering
\includegraphics[width=\textwidth]{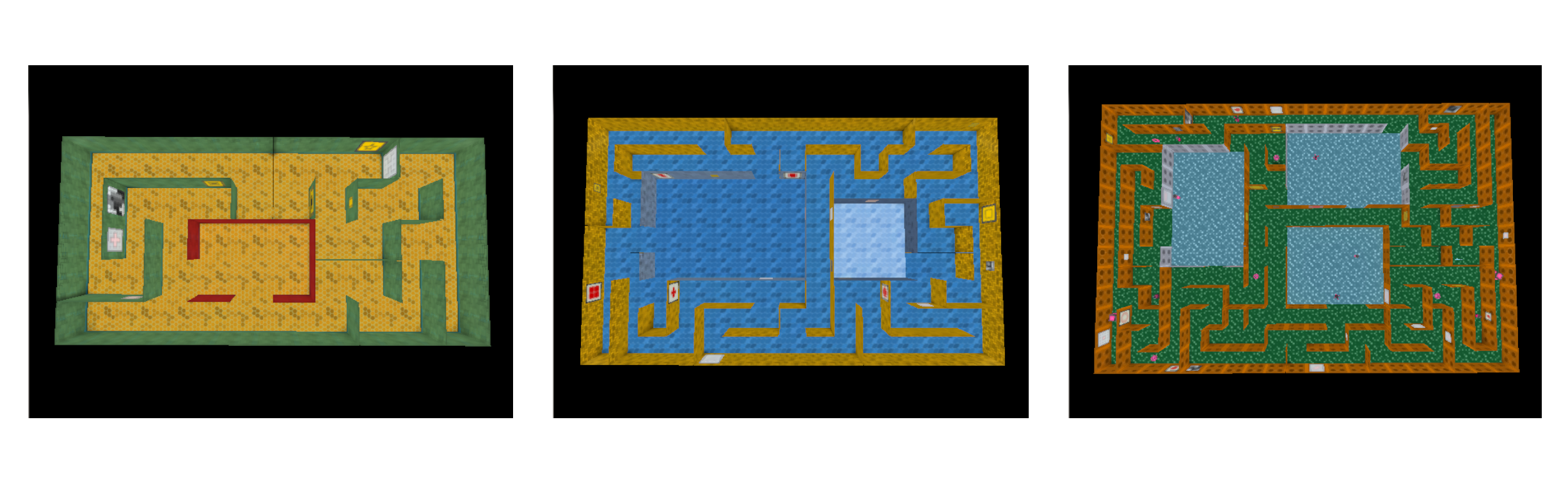}
  \caption{\small{Top-down views of static maze levels. Left: $\mathrm{nav\_maze\_static\_01}$, middle: $\mathrm{nav\_maze\_static\_02}$ and right: $\mathrm{nav\_maze\_static\_03}$}.\label{fig:mazes_top_down}}
\end{figure}

\begin{figure}[t]
\centering
\includegraphics[width = \textwidth]{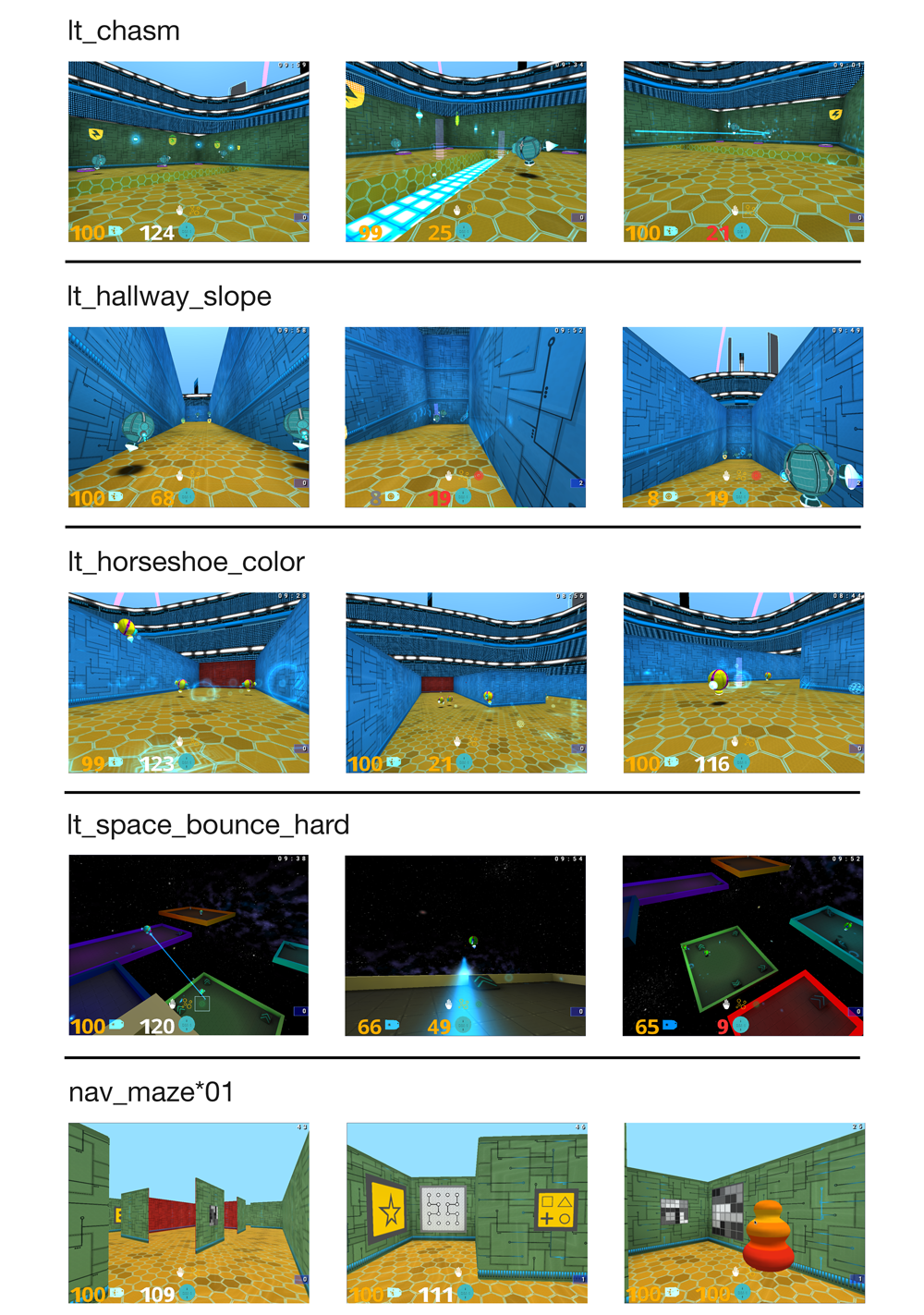}
  \caption{\small{Example images from the agent's egocentric viewpoint from several example DeepMind Lab levels.}\label{fig:first_person_views_1} }
\end{figure}

\begin{figure}[t]
\centering
\includegraphics[width = \textwidth]{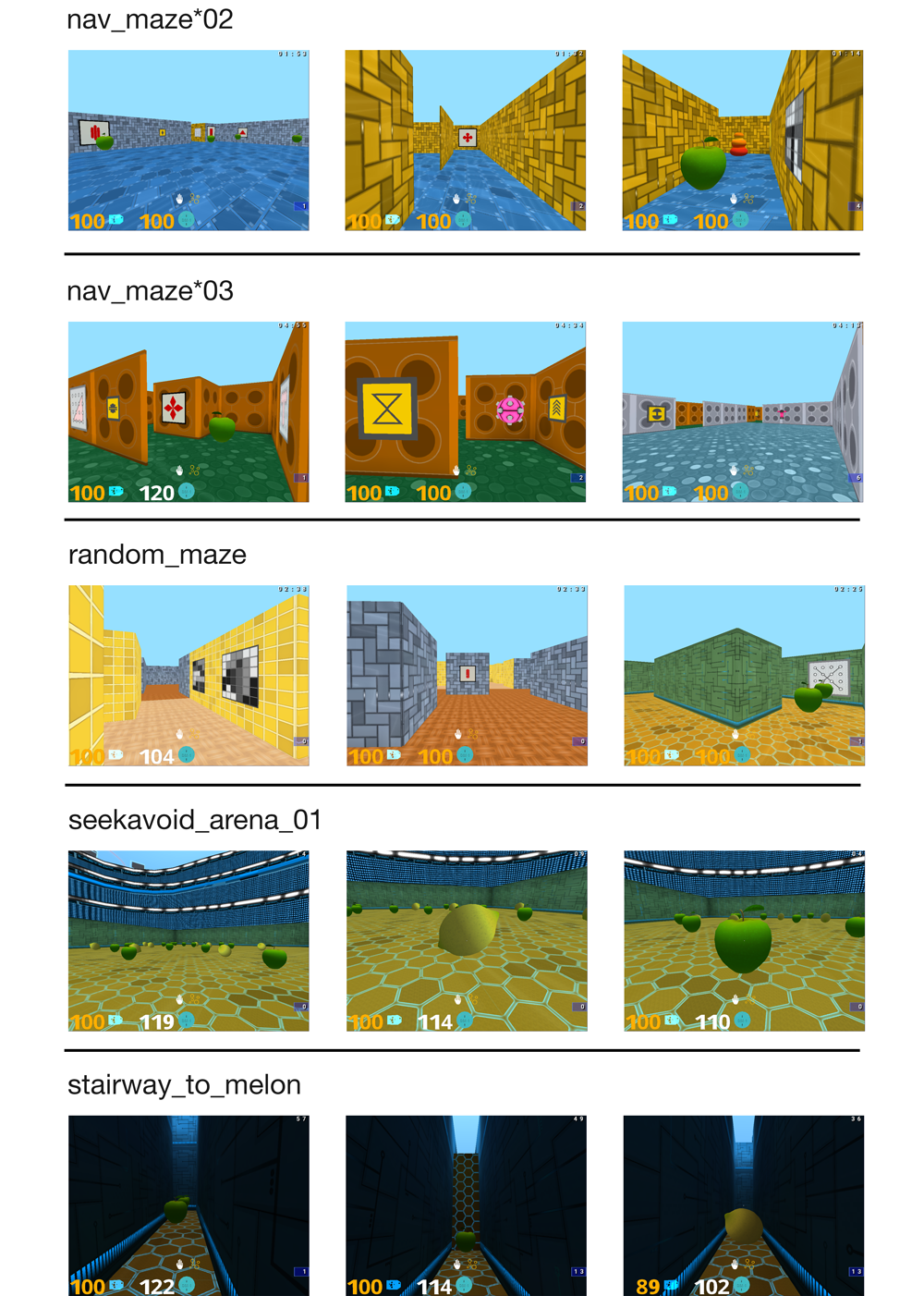}
  \caption{\small{Example images from the agent's egocentric viewpoint from several example DeepMind Lab levels.}\label{fig:first_person_views_2} }
\end{figure}

Figures \ref{fig:first_person_views_1} and \ref{fig:first_person_views_2} show a gallery of screen shots from the first-person perspective of the agent. The levels can be divided into four categories:

\begin{enumerate}
\item Simple fruit gathering levels with a static map ($\mathrm{seekavoid\_arena\_01}$ and \newline $\mathrm{stairway\_to\_melon}$). The goal of these levels is to collect apples (small positive reward) and melons (large positive reward)  while avoiding lemons (small negative reward).

\item Navigation levels with a static map layout ($\mathrm{nav\_maze\_static\_0\{1,2,3\}}$ and \newline $\mathrm{nav\_maze\_random\_goal\_0\{1,2,3\}}$). These levels test the agent's ability to find their way to a goal in a fixed maze that remains the same across episodes. The starting location is random. In the random goal variant, the location of the goal changes in every episode. The optimal policy is to find the goal's location at the start of each episode and then use long-term knowledge of the maze layout to return to it as quickly as possible from any location. The static variant is simpler in that the goal location is always fixed for all episodes and only the agent's starting location changes so the optimal policy does not require the first step of exploring to find the current goal location. The specific layouts are shown in figure \ref{fig:mazes_top_down}.

\item Procedurally-generated navigation levels requiring effective exploration of a new maze generated on-the-fly at the start of each episode ($\mathrm{random\_maze}$). These levels test the agent's ability to explore a totally new environment. The optimal policy would begin by exploring the maze to rapidly learn its layout and then exploit that knowledge to repeatedly return to the goal as many times as possible before the end of the episode (three minutes).

\item Laser-tag levels requiring agents to wield laser-like science fiction gadgets to tag bots controlled by the game's in-built AI ($\mathrm{lt\_horseshoe\_color}$, $\mathrm{lt\_chasm}$, \newline $\mathrm{lt\_hallway\_slope}$, and $\mathrm{lt\_space\_bounce\_hard}$). A reward of $1$ is delivered whenever the agent tags a bot by reducing its shield to 0. These levels approximate the usual gameplay from Quake III Arena. In $\mathrm{lt\_hallway\_slope}$ there is a sloped arena, requiring the agent to look up and down. In $\mathrm{lt\_chasm}$ and $\mathrm{lt\_space\_bounce\_hard}$ there are pits that the agent must jump over and avoid falling into. In $\mathrm{lt\_horseshoe\_color}$ and $\mathrm{lt\_space\_bounce\_hard}$, the colours and textures of the bots are randomly generated at the start of each episode. This prevents agents from relying on colour for bot detection. These levels test aspects of fine-control (for aiming), planning (to anticipate where bots are likely to move), strategy (to control key areas of the map such as gadget spawn points), and robustness to the substantial visual complexity arising from the large numbers of independently moving objects (gadget projectiles and bots). 
\end{enumerate}

\section*{Technical Details}
The original game engine is written in C and, to ensure compatibility with future changes to the engine, it has only been modified where necessary. DeepMind Lab provides a simple C API and ships with Python bindings.

The platform includes an extensive level API, written in Lua, to allow custom level creation and mechanics. This approach has resulted in a highly flexible platform with minimal changes to the original game engine.

DeepMind Lab supports Linux and has been tested on several major distributions.

\subsection*{API for agents and humans}
The engine can be run either in a window, or it can be run headless for higher performance and support for non-windowed environments like a remote terminal. Rendering uses OpenGL and can make use of either a GPU or a software renderer.

A DeepMind Lab instance is initialised with the user’s settings for level name, screen resolution and frame rate. After initialisation a simple RL-style API is followed to interact with the environment, as per figure \ref{fig:python_example}.

\begin{figure}[h]
\begin{lstlisting}[language=Python]
# Construct and start the environment.
lab = deepmind_lab.Lab('seekavoid_arena_01', ['RGB_INTERLACED'])
lab.reset()

# Create all-zeros vector for actions.
action = np.zeros([7], dtype=np.intc)

# Advance the environment 4 frames while executing the action.
reward = env.step(action, num_steps=4)

# Retrieve the observations of the environment in its new state.
obs = env.observations()  # dict of Numpy arrays
rgb_i = obs['RGB_INTERLACED']
assert rgb_i.shape == (240, 320, 3)\end{lstlisting}
\caption {Python API example.
\label{fig:python_example} }
\end{figure}

\subsection*{Level generation}
Levels for DeepMind Lab are Quake III Arena levels. They are packaged into $\mathrm{.pk3}$ files (which are ZIP files) and consist of a number of components, including level geometry, navigation information and textures.

DeepMind Lab includes tools to generate maps from $\mathrm{.map}$ files. These can be cumbersome to edit by hand, but a variety of level editors are freely available, e.g. GtkRadiant \citep{gtkradiant}. In addition to built-in and user-provided levels, the platform offers Text Levels, which are simple, human-readable text files, to specify walls, spawn points and other game mechanics as shown in the example in figure \ref{fig:text_level}. Refer to figure \ref{fig:top_down_text_level} for a render of the generated level.

\begin{figure}[h]
\begin{lstlisting}    
map = [[
**************
*  *   *******
**     *   ***
*****  I   ***
*****  *   ***
*****  *******
*****   ******
******H*******
*        I P *
**************
]]
\end{lstlisting}
\caption {Example text level specification, where ‘*’ is a wall piece, ‘P’ is a spawn point and ‘H’ and ‘I’ are doors. \label{fig:text_level} }
\end{figure}

\begin{figure}[h]
\centering
\includegraphics[scale = .35]{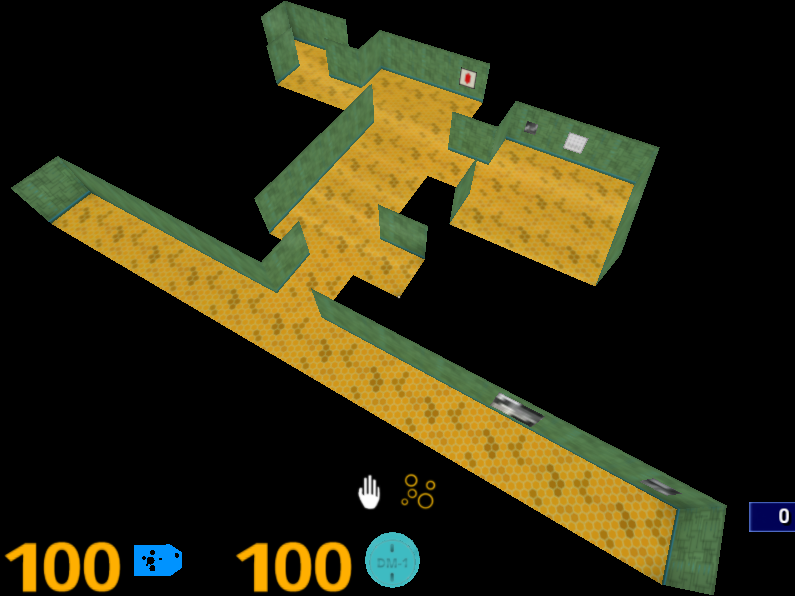}
  \caption{A level with the layout generated from the text in figure \ref{fig:text_level}. \label{fig:top_down_text_level} }
\end{figure}

In the Lua-based level API each level can be customised further with logic for bots, item pickups, custom observations, level restarts, reward schemes, in-game messages and many other aspects.     

\section*{Results and Performance}
Tables \ref{nav_maze_performance} and \ref{space_bounce_performance} show the platform's performance at different resolutions for two typical levels included with the platform. The frame rates listed were computed by connecting an agent performing random actions via the Python API. This agent has insignificant overhead so the results are dominated by engine simulation and rendering times.

The benchmarks were run on a Linux desktop with a 6-core Intel Xeon 3.50GHz CPU and an NVIDIA Quadro K600 GPU.

\begin{table}[h]
\centering
\begin{tabular}{l|ll|ll}
& \textbf{CPU} &  & \textbf{GPU} &  \\
& \textbf{RGB} & \textbf{RGBD} & \textbf{RGB} & \textbf{RGBD} \\
\hline
\textbf{84 x 84}   & 199.7 & 189.6 & 996.6 & 995.8\\
\textbf{160 x 120} & 86.8 & 85.4 & 973.2 & 989.2\\
\textbf{320 x 240} & 27.3 & 27.0 & 950.0 & 784.7
\end{tabular}
\caption{Frame rate (frames/second) on nav\_maze\_static\_01 level.}
\label{nav_maze_performance}
\end{table}

\begin{table}[h]
\centering
\begin{tabular}{l|ll|ll}
& \textbf{CPU} &  & \textbf{GPU} &  \\
& \textbf{RGB} & \textbf{RGBD} & \textbf{RGB} & \textbf{RGBD} \\
\hline
\textbf{84 x 84}   & 286.7 & 263.3 & 866.0 & 850.3\\
\textbf{160 x 120} & 237.7 & 263.6 & 903.7 & 767.9\\
\textbf{320 x 240} & 82.2 & 98.0 & 796.2 & 657.8
\end{tabular}
\caption{Frame rate (frames/second) on lt\_space\_bounce\_hard level.}
\label{space_bounce_performance}
\end{table}

Machine learning results from early versions of the DeepMind Lab platform can be found in \cite{mnih2016asynchronous, jaderberg2016reinforcement, mirowski2016learning}.

\section*{Conclusion}
DeepMind Lab enables research in a 3D world with rich science fiction visuals and game-like physics. DeepMind Lab facilitates creative task development. A wide range of environments, tasks, and intelligence tests can be built with it. We are excited to see what the research community comes up with.

\section*{Acknowledgements}
This work would not have been possible without the support of DeepMind and our many colleagues there who have helped mature the platform. In particular we would like to thank Thomas Köppe, Hado van Hasselt, Volodymyr Mnih, Dharshan Kumaran, Timothy Lillicrap, Raia Hadsell, Andrea Banino, Piotr Mirowski, Antonio Garcia, Timo Ewalds, Colin Murdoch, Chris Apps, Andreas Fidjeland, Max Jaderberg, Wojtek Czarnecki, Georg Ostrovski, Audrunas Gruslys, David Reichert, Tim Harley and Hubert Soyer.

\newpage

{\normalsize
\bibliographystyle{plainnat}
\bibliography{biblio}
}

\end{document}

%% file: macros.tex
\usepackage{amsmath, amsthm}
\usepackage{amsfonts}
\usepackage{amssymb}